\documentclass[lettersize,journal]{IEEEtran}
\usepackage{hyperref} 
\usepackage{amsmath,amssymb,amsfonts}
\usepackage{cleveref}
\usepackage[ruled,vlined]{algorithm2e}
\crefname{algocf}{algorithm}{algorithms}
\Crefname{algocf}{Algorithm}{Algorithms}
\usepackage{array}
\usepackage[caption=false,font=normalsize,labelfont=sf,textfont=sf]{subfig}
\usepackage{textcomp}
\usepackage{stfloats}
\usepackage{url}
\usepackage{verbatim}
\usepackage[table]{xcolor}
\usepackage{graphicx}
\usepackage{cite}
\usepackage{comment}
\usepackage{balance}
\usepackage{orcidlink}
\usepackage{tikz}
\begin{document}

\newcommand\submittedtext{%
  \footnotesize This work has been submitted to the IEEE for possible publication. Copyright may be transferred without notice, after which this version may no longer be accessible.}

\newcommand\submittednotice{%
\begin{tikzpicture}[remember picture,overlay]
\node[anchor=south,yshift=10pt] at (current page.south) {\fbox{\parbox{\dimexpr0.65\textwidth-\fboxsep-\fboxrule\relax}{\submittedtext}}};
\end{tikzpicture}%
}

\title{Adaptive Active Inference Agents for Heterogeneous and Lifelong Federated Learning
\thanks{\textsuperscript{*}Corresponding author: Anastasiya Danilenka}
}
\author{Anastasiya Danilenka~\orcidlink{0000-0002-3080-0303}, Alireza Furutanpey~\orcidlink{0000-0001-5621-7899}, Victor Casamayor Pujol~\orcidlink{0000-0003-2830-8368}, Boris Sedlak~\orcidlink{0009-0001-2365-8265}, Anna Lackinger~\orcidlink{0009-0006-2908-0528}, Maria Ganzha~\orcidlink{0000-0001-7714-4844}, Marcin Paprzycki~\orcidlink{0000-0002-8069-2152}, Schahram Dustdar~\orcidlink{0000-0001-6872-8821},~\IEEEmembership{Fellow,~IEEE}
\thanks{Anastasiya Danilenka and Maria Ganzha are with Faculty of Mathematics and Information Science, Warsaw University of Technology (email: anastasiya.danilenka.dokt@pw.edu.pl, maria.ganzha@pw.edu.pl)}
\thanks{Alireza Furutanpey, Boris Sedlak, Anna Lackinger, and Schahram Dustdar are with Distributed Systems Group, TU Wien (email: a.furutanpey@dsg.tuwien.ac.at, boris.sedlak@dsg.tuwien.ac.at, a.lackinger@dsg.tuwien.ac.at, dustdar@dsg.tuwien.ac.at)}%

\thanks{Victor Casamayor Pujol is with Distributed Intelligence and Systems-Engineering Lab, Universitat Pompeu Fabra (email: victor.casamayor@upf.edu)}%

\thanks{Marcin	Paprzycki is with Systems Research Institute, Polish Academy of Sciences (email: paprzyck@ibspan.waw.pl)}%

}



\maketitle
\submittednotice
\begin{abstract}
Handling heterogeneity and unpredictability are two core problems in pervasive computing. The challenge is to seamlessly integrate devices with varying computational resources in a dynamic environment to form a cohesive system that can fulfill the needs of all participants.
Existing work on adaptive systems typically focuses on optimizing individual variables or low-level Service Level Objectives (SLOs), such as constraining the usage of specific resources. While low-level control mechanisms permit fine-grained control over a system, they introduce considerable complexity, particularly in dynamic environments. 
To this end, we propose drawing from Active Inference (AIF), a neuroscientific framework for designing adaptive agents. Specifically, we introduce a conceptual agent for heterogeneous pervasive systems that permits setting global systems constraints as high-level SLOs. Instead of manually setting low-level SLOs, the system finds an equilibrium that can adapt to environmental changes. We demonstrate the viability of our AIF agents with an extensive experiment design, using heterogeneous and lifelong federated learning as an application scenario. We conduct our experiments on a physical testbed of devices with different resource types and vendor specifications.
The results provide convincing evidence that an AIF agent can adapt a system to environmental changes. In particular, the AIF agent can balance competing SLOs in resource heterogeneous environments to ensure up to 98\% fulfillment rate.
\end{abstract}


\section{introduction}
\label{sec:introduction}
The Distributed Computing Continuum is an emerging paradigm for systems that can seamlessly integrate multiple layers of computing infrastructure~\cite{dustdar_distributed_2023}.
Computing continuum systems promise to enable infrastructure-critical pervasive applications with stringent requirements, such as Mobile Augmented Reality (MAR) for cognitive applications~\cite{cogxr} and remote sensing for disaster management~\cite{edgedm}. There are three recurrent characteristics among pervasive applications deployed on a continuum. First, is their \textit{reliance on AI-based methods} for tasks that classical control structures cannot solve efficiently or with sufficient precision~\cite{eiconfluence}. For example, MAR applications must process streams of high-dimensional data that a service could ideally process at the source to fulfill a sub-10 millisecond latency Service Level Objective (SLO).
The caveat is that resources in proximity are constrained. Typical solutions involve task partitioning and lightweight data reduction methods that minimize the penalty for offloading to remote resources~\cite{frankensplit}. The second characteristic is \textit{heterogeneity}, i.e., resource-asymmetry, vendor specifications, and usage patterns. Although pervasive applications follow an overall common objective, a system must consider the individual properties and objectives of participants. Third, is the \textit{continuously drifting problem domain} intrinsic to the dynamic environments of pervasive applications, such that the source distribution drifts over time and data volume is non-static. 
Conclusively, a necessary precondition is a system that can adapt to non-identically and independently distributed (non-IID) data. Moreover, the system must facilitate collaboration between heterogeneous devices to fulfill their SLOs by distributing workload fairly and considering the individual properties of participants. 

The focus of this paper is on \textit{lifelong heterogeneous federated learning} (FL) as we find it best encapsulates the primary challenges of pervasive applications that share the described characteristics. In general, FL participants collaborate for a common objective, i.e., to maximize the prediction performance. Yet, each participant has a private local validation set to determine whether their criteria are locally met. Time constraints that ensure smooth operations and resource asymmetry further instigate friction when attempting to satisfy local objectives. Hence, despite a common objective, to fulfill the SLOs of each participant individually, a delicate balance is necessary. Lastly, the dynamic environment gradually drifts the distribution and varies the data volume. 

This work aims to demonstrate the viability of \textit{Active Inference (AIF)} in designing adaptive agents that can gracefully handle the challenging requirements of pervasive applications. While AIF is a neuroscientific framework, recent work has shown promising results by conceiving methods from the underlying ideas for workload scheduling in distributed systems~\cite{sedlak_equilibrium_2024,sedlak_adaptive_2024}. In particular, we find that the objectives of Active Inference and pervasive computing intrinsically intertwine. Context awareness is crucial for pervasive applications as these systems operate in dynamic environments and must adapt to changes in their surroundings. Precisely context awareness is a defining characteristic of AIF agents. However, the current application of AIF for systems is more conceptual and only partially implements the core components of the AIF framework. Other work on systems that adapt to changing requirements typically focuses on optimizing individual variables, such as learning rate or setting low-level SLOs as constraints on specific resources~\cite{shubha_adainf_2023, KUNDROO2023101740}. Despite providing more fine-grained control, it is unreasonable to expect application developers to understand the implications of each low-level constraint to the overall system, particularly in dynamic environments where resources are scarce and availability is less predictable. In contrast, our AIF agent permits setting high-level SLO targets to find an equilibrium, defined as the system configuration that fulfills all its SLOs, without attempting to enforce constraints from possibly conflicting low-level SLOs. 

We design experiments that accurately reflect the relevant real-world conditions by implementing a physical testbed consisting of heterogeneous devices with varying resource types and computational capabilities. Additionally, we leverage a controlled process for data generation to evaluate the adaptability to a dynamic environment precisely. We extensively evaluate our agents with a strong emphasis on reproducibility. The results underpin the claim that an AIF agent can successfully balance competing SLOs among clients despite considerable resource asymmetry and adapt to the dynamic environment. Still, we transparently discuss current limitations by accentuating the parts of our result that best show our agent’s weaknesses. The intention is to foster research interest in AIF from a systems perspective, as we sincerely believe that it poses an exceptionally promising research direction for pervasive applications and the compute continuum. Naturally, we open-source our repository as an addition to the community to reproduce, scrutinize, and extend our approach~\footnote{\url{https://github.com/adanilenka/adaptive_aif_agents_for_fl}}.

We summarize our contributions as:
\begin{itemize} 
   \item An adaptive mechanism for heterogeneous lifelong FL based on AIF which allows handling non-IID data distributions and heterogeneous device characteristics inherent in pervasive computing environments.
    \item A conceptual AIF agent that balances multiple SLOs during model training. When SLOs have competing targets, agents can autonomously infer optimal training configurations without manual intervention.
    \item Empirical evaluation of AIF agents for pervasive FL tasks under real-world-inspired conditions, incorporating data and resource heterogeneity through a reproducible experimental setup with a physical testbed of heterogeneous devices.
\end{itemize}
\section{Background and Related Work} \label{sec:background}

\subsection{Lifelong Heterogeneous Federated Learning}  \label{subsec:bgfl}

In Federated Learning, participants train a global model to maximize prediction performance without disclosing private data. Participants optimize and validate the model parameters with their local dataset in a \textit{training round} before aggregating their weights globally.

\subsubsection{Lifelong Federated Learning}
 FL is \textit{lifelong} when training continuously adapts to concept drifts and other changes occurring in continuous data streams
 ~\cite{SUAREZCETRULO2023118934}. Introducing concept drifts is an intrinsic property of the dynamic deployment environment of pervasive applications. The presence of concept drift can lead to both prolonged time until model convergence and reduced model performance, stressing the importance of treating concept drifts in FL and the need for targeted solutions for different types of concept drifts~\cite{10415779}. Yet, there is limited research in lifelong learning for FL~\cite{pmlr-v206-jothimurugesan23a}. The current approaches to adapt to concept drifts rely on custom drift detectors to understand when the drift occurs~\cite{jothimurugesan2023federated, 10839814}, which results in the need to tune said detectors to the use case and potential drift scenarios, leaving the challenge on how to distinguish drifts from anomalous data. Moreover, although concept drifts are one of the most challenging issues to face for lifelong FL, they are not the only one, as discussed further in this section. This highlights the need for more self-adapting mechanisms, e.g., by introducing meta-learning concepts to FL~\cite{10468591}, that can maintain the performance of the FL systems.

\subsubsection{Heterogeneous Federated Learning}
We refer to Federated Learning as heterogeneous when data is non-IID and hardware specifications vary among participants. Moreover, hardware heterogeneity typically implies that resources are constrained, as less powerful devices, often located closer to the data source, must also be accommodated in the learning process. 

Optimizing FL workflows, especially in the presence of heterogeneities, is important for minimizing the time-to-accuracy of model training~\cite{jiang2022towards}.

In that context, Kundroo et al.~\cite{KUNDROO2023101740} proposed \textit{FedHPO}, a federated optimization algorithm that accelerates each client's training by modifying its hyperparameters, such as learning rate or epochs. However, FedHPO introduces additional algorithm parameters to set and tune, e.g., patience or thresholds to guide the optimization process, which limits its flexibility in dynamic environment usage. To optimize for local training time in dynamic and heterogeneous devices conditions, an asynchronous FL approach \textit{FedTS} was proposed by Li et al.~\cite{10540355}, empowering the FL server to detect and optimize for slower clients. Still, the scheme focuses on ensuring time constraints for heterogeneous devices and does not cover lifelong scenarios with dynamic data distributions.

Several studies also explored multi-objective optimization (MOO) in FL to balance competing objectives. One approach is to optimize neural network models instead of client-specific hyperparameter optimization \cite{zhu2019multi,paragliola2022evaluation}.
Additionally, a significant number of existing research focuses on optimizing client selection or clustering instead of adjusting the parameters of individual clients to reach multi-dimensional goals \cite{Badar_Sikdar_Nejdl_Fisichella_2024, lackinger2024inference, shuai2023node}. 
Although these approaches offer practical solutions for balanced MOO, they do not take into account the individual parameters of heterogeneous clients.
Moreover, a vast area of research lies in adopting Bayesian Optimization (BO)~\cite{NIPS2012_05311655} to the needs of FL. Along with grid search, BO is used for hyperparameter tuning in FL~\cite{NEURIPS2020_6dfe08ed}, still, the downside of both methods includes their inherent incentive to find the best possible hyperparameters set (either in one-shot fashion or during multiple communication rounds) which makes it more difficult to adapt them to changing conditions of lifelong learning. Another drawback of the classical BO is its focus on one singular objective to optimize, e.g., model accuracy. This problem is starting to be addressed by Multi-objective Bayesian Optimization (MOBO)~\cite{Badar_Sikdar_Nejdl_Fisichella_2024}. Yet, the presented MOBO approach also
does not consider lifelong learning scenarios.

A number of automated optimization tools are available to use for hyperparameter optimization tasks and were adopted by the industry, such as HyperOpt~\cite{pmlr-v28-bergstra13} and Optuna~\cite{10.1145/3292500.3330701}, with such platforms as RayTune~\cite{liaw2018tuneresearchplatformdistributed} allowing for distributed parameter tuning. Lately, FL research started to adopt hyperparameter optimization tools~\cite{10.1145/3605098.3636015, krouka2024communicationefficient, 10679103, 10.14778/3617838.3617842, 10.1145/3555776.3577847}. However, currently, the application of hyperparameter tuning still does not cover lifelong learning scenarios. 

Therefore, existing work on optimization in FL does not consider changing environments and lifelong FL scenarios or lacks individual clients' hyperparameter tuning in general, which is crucial for pervasive applications.

\subsubsection{Service Level Objectives for Federated Learning}
SLOs are definable constraints on a system that operators may use as contracts with application developers~\cite{nastic2020SLOC, casamayor2024invited}. Low-level SLOs quantify directly observable measures, such as CPU or memory usage. High-level SLOs abstract low-level SLOs to reduce the difficulty of diagnosing and configuring complex and wide-spanning systems, i.e., compute continuums with measures such as throughput or monetary costs. 

For our purposes, high-level SLOs provide an intuitive interface to set targets for an AIF agent and quantitative measures to determine its adaptability to a dynamic environment. In particular, maintaining prediction performance and minimizing round duration are two primary objectives for lifelong heterogeneous federated learning. An SLO on prediction performance ensures consistent solution quality. In contrast, an SLO on timeliness is crucial as resources are constrained, and a considerably slower client can delay global weight updates. Time and prediction performance SLOs abstract more detailed system parameters that have an impact on them, focusing on end-user experience and overall system performance.

There exist multiple approaches that aim to combine SLOs with dynamic processing requirements: Zhang et al.~\cite{zhang_octopus_2023} presented \textit{Octopus} -- the framework that finds optimal services configurations in multi-tenant edge computing scenarios. \textit{Octopus} predicts SLO fulfillment of two variables based on a deep neural network. 
Shubha et al.~\cite{shubha_adainf_2023} presented \textit{AdaInf}, which detects SLO violations of a GPU scheduling task whenever variable drifts occur. Through \textit{AdaInf}, it is possible to find SLO-fulfilling resource allocations between model training and inference. Although these approaches are SLO-aware, oriented at continuous processes, and may utilize agents for the decision-making process, they are primarily used for inference services and do not consider the scenario of optimizing the training of ML models, which require more flexible and high-level SLOs definitions, appropriate for the considered FL/ML scenario.

\subsection{Active Inference}
\label{subsec:bgaci}
Active inference is a neuroscientific framework based on the free energy principle (FEP)~\cite{friston2006free}. AIF agents adjust their model according to new observations and enact environmental changes to suit their preferences. The objective is to minimize the difference between the agent's internal representation and real-world models, i.e., to adapt to its environment. In principle, the underlying framework of AIF generally applies to adaptive systems~\cite{friston_designing_2024}. Therefore, it is reasonable to assume that AIF is a promising direction for computing continuums that must adapt to a dynamic environment~\cite{sedlak_equilibrium_2024}. 

AIF agents continuously evaluate the expected free energy (EFE) for different policies and assess their impact on underlying models. EFE constitutes the planning ability of AIF agents, as it allows for evaluating policies of custom length into the future, utilizing the current understanding of the world model (generative world model) as the source for simulations. In a system's context, an agent who understands the world model will minimize EFE by selecting policies likely to fulfill SLOs.
We describe EFE with two distinct components~\cite{parr2022active}:
\begin{equation}
\label{eq:efe}
    \text{EFE} = - \overbrace{\mathbb{E}_{Q(o|\pi)} [\ln P(o|C)]}^{\text{Pragmatic Value}} -  \underbrace{\mathbb{E}_{Q(o,s|\pi)}D_{\mathrm{KL}} \left[ Q(s|o) \, \| \, Q(s) \right]}_{\text{Information Gain}}
\end{equation}

The \textit{information gain} (IG) estimates how much the model can improve by choosing a particular policy and aims to resolve the uncertainty currently present in the generative world model. Thus, IG takes into consideration the approximate posterior Q over hidden states (variables in the world that the agent cannot observe) given observations and Q of hidden states only -- our prior beliefs about the model before any observation. Here, the agent aims at \textit{maximizing} the divergence, thus, looking for the most informative future steps. Conversely, the \textit{pragmatic value} (PV) estimates how close a possible outcome  (observation O) will be to the agent’s preferred observations (C), focusing on meeting the expectation defined for the agent. Together, IG and PV balance the exploration/exploitation trade-off.

The cases in which AIF was used to dynamically support computing systems are mainly focused on robotics; Oliver et al.~\cite{oliver_empirical_2022} give a comprehensive overview of how AIF allows (robotic) systems to act under uncertainty. Nevertheless, the application of AIF extends to continuous stream processing systems, such as provided by Sedlak et al.~\cite{sedlak_equilibrium_2024,sedlak_adaptive_2024}, which uses a wide set of processing metrics as sensory observations. 
Actions taken by the processing system were elastic adaptations, e.g., scaling resources or quality, allowing to empirically find system configurations that fulfill SLOs.

Proved useful for ensuring the adaptability of robotic and stream processing systems, AIF could address the existing research gap in optimizing dynamic pervasive FL systems.

To sum up, while AdaInf and Octopus are SLO-aware inference services, they are designed for dynamic and resource-efficient \textit{model serving}, particularly in scenarios like multi-model and edge inference. In contrast, traditional FL hyperparameter tuning methods, such as BO and grid search, aim to achieve one best FL model but do not inherently address the adaptability and continuity required for lifelong learning. Active Inference, with its inherent adaptability, innate exploration-exploitation trade-off handling, and potential explainability through causality, offers a promising theoretical solution for the presented gap. To the best of our knowledge, this paper is the first attempt to apply AIF to FL (and, by extension, lifelong FL), addressing the challenges of adaptability and heterogeneity in this domain.
\section{Problem Statement} \label{sec:probstat}
\begin{figure}[htbp]
\centerline{\includegraphics[width=\columnwidth]{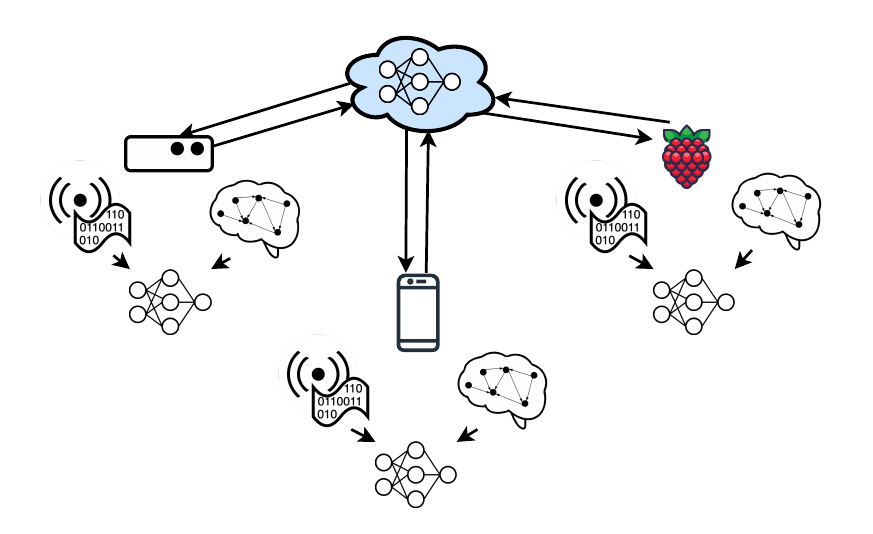}}
\caption{Heterogeneous FL with data streams and AIF agents}
\label{fig:fl_scheme}
\end{figure}

We consider a lifelong heterogeneous FL system consisting of an orchestrator with $N$ participants. In \Cref{fig:fl_scheme} we can observe a central model and orchestrator in the Cloud that communicates with different IoT/Edge devices. These produce a stream of data that is used by an ML model; at the same time, these devices embed an AIF agent that adjusts the training of their ML models to achieve optimal (SLO fulfillment) performance.  

The objective of the FL system described in  \Cref{fig:fl_scheme} is to maximize the overall SLO fulfillment across all timestamps. The challenge is to ensure the highest possible SLO fulfillment (reaching equilibrium), given heterogeneous participants and data within a dynamic environment. Client hardware is heterogeneous in vendor specification, available resource types, and overall computational capacity. For example, some devices may have onboard accelerators, such as GPUs, and others may only work with energy-efficient CPUs. 
The data source is non-IID with temporal correlations, i.e., the training must adapt to non-stationary data. 
SLOs are set on a global system level and all clients aim to fulfill the same SLOs. SLOs aim to ensure smooth operations, i.e., timely training and consistently adequate model performance, and clients check after each training round whether the SLOs are fulfilled locally. 

We introduce a mechanism to control and manage SLO fulfillment by defining FL training \textit{configurations}, which specify training parameters that directly affect the system’s ability to fulfill SLOs.  Configurations function as levers that an agent can change to fulfill the high-level SLOs.
\section{Proposed Method} \label{sec:method}
\begin{figure}[htbp]
\centerline{\includegraphics[width=\columnwidth]{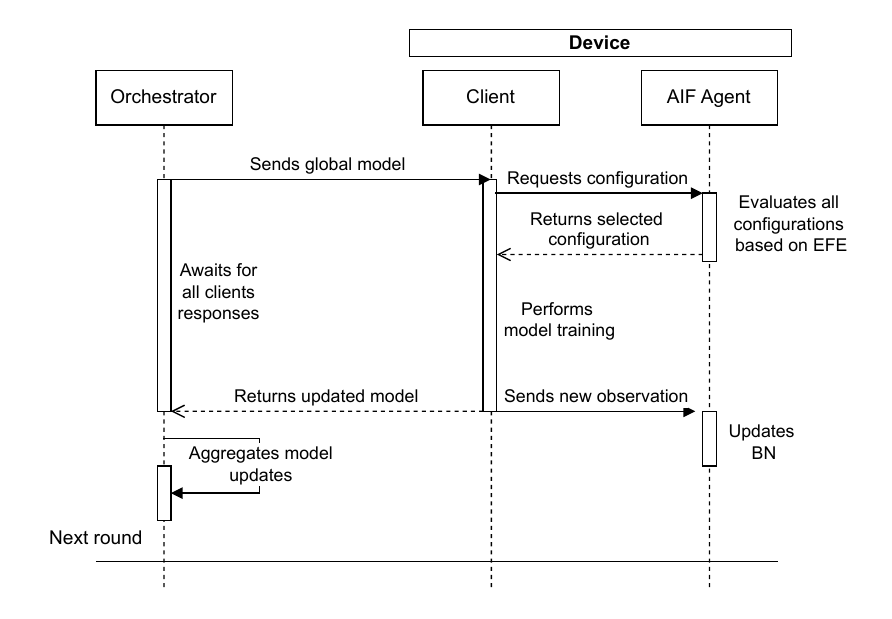}}
\caption{Sequence diagram for one FL round of the proposed method}
\label{fig:sequence_diagram}
\end{figure}

This section presents the design of the AIF agents that optimize a heterogeneous and lifelong federated system to fulfill SLOs. \Cref{fig:sequence_diagram} illustrates how system entities interact. (1) The orchestrator sends the current global model to the client, initiating a new FL round; (2) the client requests a configuration from the AIF Agent; (3) the AIF agent evaluates configurations based on EFE and returns the best one; (4) the client trains the model using the selected configuration; (5) after training, the client sends a new observation to the AIF Agent and returns the updated model to the Orchestrator, which aggregates all updates; (6) the AIF agent updates its world model (Bayesian Network) for future configuration optimization; (7) the next FL round begins. 
\Cref{alg:train} summarizes the overall procedure for client-side training with the rest of the section elaborating on notable details about the process for an agent to find and choose optimal FL training configurations in a dynamic environment.
\begin{algorithm}[h]
\caption{On Client Training Procedure}
\label{alg:train}

\SetKwProg{Fn}{Procedure}{}{}

\Fn{\textsc{train}($global\_model$, $is\_lifelong$)}{
    $train\_set$ $\gets$ \textsc{fetch\_next\_train\_set}() \\
    $config$, $expected\_ig$ $\gets$ \textsc{infer\_best\_config}() \\

    \textbf{With} $config$: \\
    \Indp $updated\_model$, $metrics$ $\gets$ \textsc{train\_model}($global\_model$, $train\_set$, $config$) 

    \Indm 

    $slos\_fulfilled$ $\gets$ \textsc{check\_slo\_fulfillment}() \\
    \textbf{If} $is\_lifelong$: \\ 
    \Indp 
    $new\_obs$ $\gets$ $slos\_fulfilled$ $\cup$ $config$ $\cup$ $metrics$ \\

    \textsc{update\_bn}($new\_obs$, $expected\_ig$) \\
    \Indm
    \KwRet{$updated\_model$, $metrics$}
}

\vspace{10pt}

\Fn{\textsc{infer\_best\_config()}}{
    $configs \gets \{\}$ \\
    \ForEach{$c$ \textbf{in} $possible\_configs$}{
        $EFE_{c}$, $ig_{c}$ $\gets$ \textsc{calculate\_efe}($c$) \\
        $configs \gets configs \cup (EFE_{c}, ig_{c})$
    }
    \KwRet{$possible\_configs_{\arg \min (\text{configs.EFE})}$}
}

\vspace{10pt}

\Fn{\textsc{update\_bn}($new\_obs$, $expected\_ig$)}{
    $obs\_surprise$ $\gets$ \textsc{calculate\_surprise}($BN$, $new\_obs$) \\
    \eIf{$obs\_surprise > expected\_ig$}{
        $BN \gets$ \textsc{do\_structure\_learning}() \\
        $BN \gets$ \textsc{do\_parameter\_estimation}($BN$)
    }{
        $BN \gets$ \textsc{do\_parameter\_update}($BN$, $new\_obs$)
    }
}

\end{algorithm}
\subsection{Learning a Simple World Model}
\begin{figure}[htbp]
\centerline{\includegraphics[width=\columnwidth]{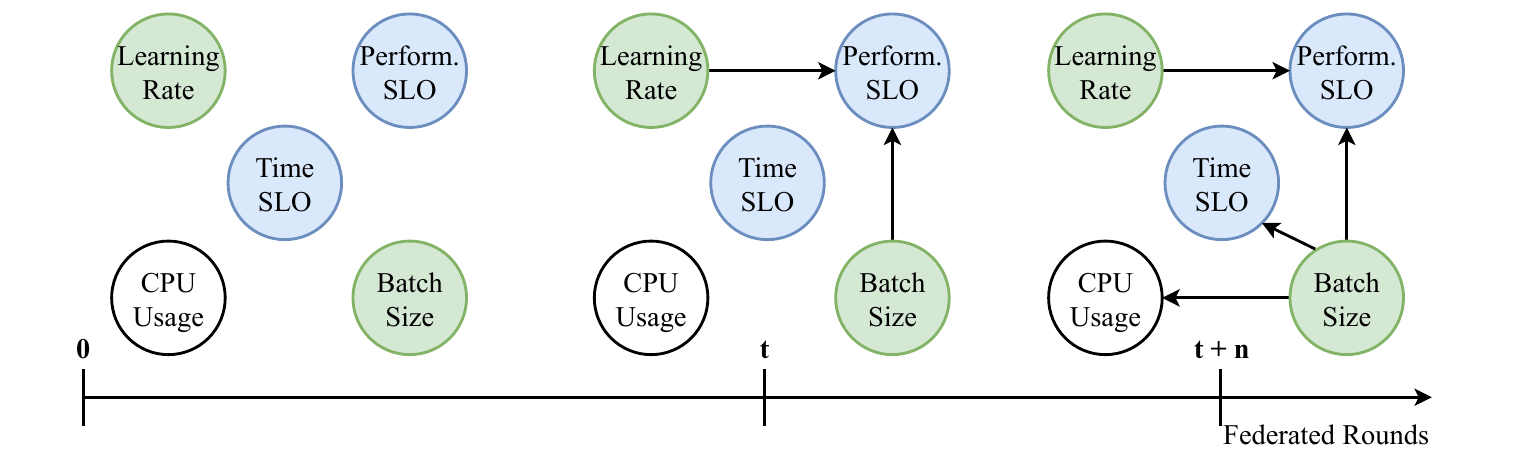}}
\caption{BN structure update throughout FL training rounds (blue vertices represent SLOs, green -- configuration variables)}
\label{fig:BN_update}
\end{figure}

The \textit{generative model} is at the core of an AIF agent, i.e., as an agent interacts with its environment, it updates its internal world representation in a perception-action cycle to improve its understanding and align its behavior to reach set goals. We choose Bayesian Networks (BNs) as they provide interpretable graphical representations of learned causal structures and offer a principled framework for probabilistic reasoning. This work considers discrete BNs with uniform priors.
Each agent's initial Bayesian network structure is unknown, as there are no assumptions on prior knowledge of the environmental dynamic. The agents require only starting knowledge of the BN variables and their respective cardinalities, thus, specifying the considered features in the environment and their respective precision. We define the BN  \( \mathcal{B} \)  of an agent  as:
\[
\mathcal{B} = (\mathcal{G}, \mathcal{P})
\]
where \( \mathcal{G} = (V, E) \) is a directed acyclic graph (DAG), and \( \mathcal{P} \) is the set of conditional probability distributions:
\[
\mathcal{P} = \{ P(X_i | \text{Pa}(X_i)) \}_{i=1}^{n}
\]
\( \text{Pa}(X_i) \) represents the parents of \( X_i \) in \( \mathcal{G} \), for which the joint distribution of the variables is factorized as:
\[
P(X_1, X_2, \dots, X_n) = \prod_{i=1}^{n} P(X_i | \text{Pa}(X_i))
\]
The BN vertices are divided into three categories:
\begin{enumerate}
\item \textbf{Configuration vertices}: Represent the (hyper)parameters of the system that are available for the agent to set.

\item \textbf{SLO vertices}:  Binary vertices that encode SLO being fulfilled or not and allow for finding dependencies between SLOs (and their fulfillment) and other vertices of the BN.

\item \textbf{System vertices:} Additional vertices provide a more comprehensive overview of the system dynamic, such as resource usage, and their connection to SLOs.
\end{enumerate}
\Cref{fig:BN_update} illustrates the process of learning the structure of the BN as the FL rounds progress. We use Hill Climb Search~\cite{10.5555/1795555} and Bayesian estimation to perform structural learning and parameter estimation.  We use variable estimation to perform exact inference, such that an AIF agent utilizes precise computation to leverage the uncertainty of BNs. As the FL rounds progress, the BN causal structures are progressively discovered. Moreover, the agent experiences further refine the conditional probability distributions.

To allow the agent to adapt to significant discrepancies between expected and observed outcomes, we distinguish between two update types. If the observed IG is higher than expected, the agent discards structure information from the previous iterations and initializes structure re-learning of the BN. Structure re-learning prioritizes the edges that include SLOs as the dependent node, to ensure that the relations that have an immediate impact on agent decisions are considered first. Conversely, if the observed IG is within expectations (equal or below expected IG), the agent only initializes a parameter update on the BN, merely accommodating new data that followed the expectations of the current generative model.

To ensure the BN does not learn from early-stage ML model performance data, that do not accurately describe the relationship between performance SLO and configuration, and instead focuses on the \textit{lifelong} part of the training only, observations from early FL rounds are omitted until the model performance stabilizes. When the global FL model performance gets sufficiently close to the target performance SLO, a lifelong learning flag signals the AIF agent to start learning, before that, the configuration is taken at random, as no pre-defined strategies are defined for this warm-up stage.

\subsection{EFE and SLO-aware Configuration Selection}
Due to SLO and configuration vertices present in the BN, the task is to choose the configuration with the highest chance of fulfilling SLOs by correctly discovering the connections between the vertices and resolving uncertainty about the world. 
The agent calculates the EFE for each configuration available to the system to determine configuration optimality using the formula in \Cref{eq:efe}. Specifically, it calculates the pragmatic value as:
\begin{equation}
\begin{aligned}
\text{Pragmatic Value} &= \sum_{\text{SLOs}} P(\text{SLOs} \mid \text{configuration}) \\
&\times \text{preference vector}
\end{aligned}
\end{equation}
and the Information Gain as:
\begin{equation} \label{eq:expected_ig}
\text{Information Gain} = I(\mathcal{A}, \mathbf{q})
\end{equation}

The preference vector encodes the agent’s goal, i.e., the desired outcome, and is expressed as a logarithm of the normalized preferences. The list of outcomes consists of all combinations of possible SLO values. Since the agent is to fulfill all set SLOs, the state where all SLOs are fulfilled will have a higher preference. For example, for one binary outcome SLO, one can set the preference vector to [0.001, 0.999], specifying that the second outcome is preferred. 

Information gain quantifies the expected Bayesian surprise that measures how much observing new data would update the agent's belief about hidden states. First, for each possible configuration, the agent assesses which observation is most likely. An \textit{observation} is a configuration and an associated outcome. The agent uses a particular configuration as evidence to predict the possible observation with a maximum a posteriori (MAP) query to simulate possible future. Then, following the implementation in \cite{Heins2022}, the calculation (Equation~\ref{eq:expected_ig}) uses the likelihood of SLOs fulfillment (matrix $A$) and the predictive density over hidden states $q$ derived from the BN to predict IG of a specific configuration. 

In summary, the information gain and pragmatic value balance a trade-off between exploring and taking the actions that most likely result in SLOs fulfillment. Once the agent has selected a configuration, the local training round starts. On completion, the agent collects lower-level metrics and checks SLO fulfillment. Lastly, it associates the outcomes with the configuration and adds it to the history dataset as a new observation for further updates.

\section{Evaluation} \label{sec:eval}
\subsection{Experiment Design}
The experiment design examines the AIF agent’s behavior and adaptability to heterogeneity and lifelong FL. 
\subsubsection{Test Bed}
We implement a physical testbed with constrained devices to replicate a heterogeneous resource environment. Additionally, we use a virtual machine with server-grade hardware for experiments in more controlled environments. \Cref{tab:hwconf} summarizes the hardware specifications.
\begin{table}[htb]
\centering
\caption{Testbed Hardware Specifications}
\label{tab:hwconf}
\resizebox{\columnwidth}{!}{%
\begin{tabular}{ccc}
\hline
Device          & CPU                 & Accelerator        \\ \hline
Virtual Machine & 8x Xeon @ 3.7 GHz   & Tes. 2560 CC       \\
\rowcolor[HTML]{EDEDED} 
Orin NX         & 8x Cortex @ 2 GHz   & Amp. 1024 CC 32 TC \\
Xavier NX       & 4x Cortex @ 2 GHz   & Vol. 384 CC 48 TC  \\
\rowcolor[HTML]{EDEDED} 
Raspberry Pi 4   & 4x Cortex @ 1.8 GHz & N/A                \\
Raspberry Pi 5   & 6x Cortex @ 2.0 GHz & N/A               
\end{tabular}%
}
\end{table}
\subsubsection{Implementation Details}
We implement the prediction model as a simple Artificial Neural Network (ANN) with PyTorch consisting of two fully connected layers using ReLU activation for non-linearity. 

We extend the Flower~\cite{beutel2020flower} framework to support FL. We implement the agent BNs with pgmpy~\cite{Ankan2015} and information gain with pymdp~\cite{Heins2022}. We implement a controllable data generation process using River~\cite{10.5555/3546258.3546368}. A more detailed technical description is out of scope and we refer interested readers to the accompanying repository.  
\subsubsection{Application Scenario}
We emulate a dynamic environment by controlling the data generation process to introduce concept and volume drifts. Each client device represents a different participant. The challenge is, for the system to adapt to the drifts or to the varying computational resources of participants. 
The experiments consider the fulfillment of two binary high-level SLOs:
\begin{enumerate}
	\item \textbf{Time}: fulfilled if a local training round does not exceed a set limit (e.g., 2 seconds).
	\item \textbf{Prediction Performance}: fulfilled if the primary validation metric (accuracy) exceeds a set value. 
\end{enumerate}
We choose time and performance as SLOs as balancing them is non-trivial. For example, focusing exclusively on fulfilling prediction performance may require spending an excessive amount of time and vice versa. The configurable hyperparameters are \textit{Batch Size} $BS \in \{8, 32, 64, 256, 512\}$ and \textit{Learning Rate} $LR \in \{0.0005, 0.001, 0.005, 0.01\}$, as there is a clear connection to them and the system’s training objective and considered drift types, e.g., learning rate tuning was proposed to battle concept drift~\cite{pmlr-v206-jothimurugesan23a}. The values chosen were selected to span across reasonable ranges that can make a difference in terms of SLOs while being distinct from each other.
Each client initializes an independent data stream locally.

The data generator $G$ is also parametrized by a drift parameter, $drift$, which controls the presence and speed of data drifts, where $drift = 0$ indicates no data drift.

In each federated round, clients train and validate their prediction model using the data samples available in an online learning fashion. At round $t$, each client $n$ possesses two datasets: $\text{Validation}_{t} = \{(x_b, y_b)\}_{b=1}^{B_t} \sim G_n$ and $\text{Train}_{t} = \text{Validation}_{t-1}$,  where $x$ is a feature vector, $y$ label assigned to the data sample and $B_t$  the size of the data set drawn from the data generator at round $t$.

This way, clients acquire a new batch of data for validation while the previous batch is re-used for training. Previous round training samples are discarded. 

\subsubsection{Baselines}


We choose two baselines to present AIF agents. The first set focuses on the presentation of the behaviour of AIF agents, aiming at identifying both adaptable behaviour pattern and assessing agents' performance. The second baseline compares AIF agents to the state-of-the-art Optuna framework to assess the optimality of hyperparameter choices made by the agents.

We define two baselines to illustrate AIF agents behaviour under data distribution drifts:
\begin{enumerate}
	\item \textbf{Random}: Represents a complete lack of adaptability and intelligent choice of hyperparameters. This baseline randomly chooses a new configuration for each federated training round.
	\item \textbf{Fixed optimal}: Represents the case where parameter tuning was performed and the optimal configuration is set once at the beginning of the training and is never changed. To select this configuration, each of the possible configurations was tested and the one with the highest mean SLOs fulfillment at the end of the observed period (around round 50) was taken as the optimal.
\end{enumerate}

To align Optuna with the AIF agent, its hyperparameter search process was modified. First, the hyperparameter study was set to cover all FL rounds, with the first hyperparameters set trial starting after the global model performance reached sufficient accuracy (same as for AIF agent as described in Section~\ref{sec:method}) and performing one trial per FL round until the end of the training. To select the configuration for the current FL round, first, an Optuna trial is performed and added to the study (stored in the local Optuna database). Then, the best configuration is inferred, and the FL training round starts. To maintain the integrity of Optuna trials in the lifelong learning process, each trial is conducted using a copy of the global model from the \textit{previous} round, evaluated on the training data from that same round. By isolating trials from ongoing learning dynamics, we prevent information leakage in both directions -- from the current global model to the trials and vice versa -- preserving the validity of the optimization process.

To make Optuna SLO-aware, the study was designed in MOO fashion, where SLOs variables served as objectives. Here, training time and 1 -- validation set accuracy served as two objectives for optimization (minimization). By default, for MOO Optuna returns not a single best configuration, but a Pareto front -- a set of feasible configurations, leaving the final choice up to the user. In the experiments, we first attempt to filter out only those configurations in the Pareto front that fulfill both SLOs, then both time and performance metrics are normalized and weighted equally into one feature, where the best configuration is returned for the FL round training. 

Each experiment describes the results regarding SLO fulfillment, as the evaluation metric expected by the users, and EFE dynamics that explain the learning and adaptation of AIF agents. 

Cumulative SLO fulfillment at round $t$ is calculated as:

\begin{equation}
\text{SLO Fulfillment}_{t} = \frac{\sum_{i=1}^{t} \text{SLO fulfilled}_i}{t}
\end{equation}

Each experiment was repeated ten times with different random seeds that controlled the random processes, such as the data generator and ANN weights initialization. For the evaluation, the first considered timestep included in SLO fulfillment tracking for a particular client run was the one where the ``lifelong'' flag (model performance stabilizes and the agent starts learning)  becomes true. The reported results are averaged across all participating clients and experiments, if not stated otherwise. The number of local epochs was set to 3 for all experiments with no client subsampling, i.e., all FL clients participated in every round. 
The SGD optimizer was used by default. It is also worth reminding that agents learn their BNs from scratch in every experiment.

\subsection{Agent Demonstration}

\begin{figure}[htbp]
\centerline{\includegraphics[width=\columnwidth]{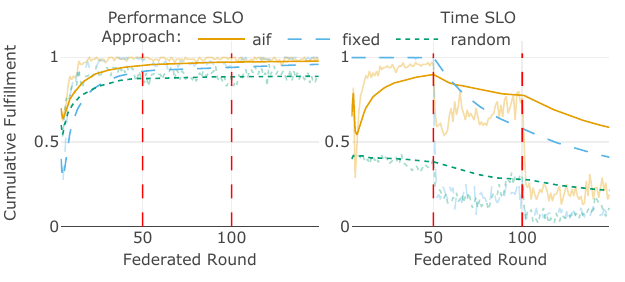}}
\caption{Mean cumulative SLOs fulfillment with two quantity drifts (red lines mark the drift start). Semitransparent lines show mean SLO fulfillment at a single federated round.}
\label{fig:slos_10_clients_quantity_skew}
\end{figure}

Consider a real-life situation where clients experience changes in the amount of collected data, e.g., seasonal demands in shops specializing in certain types of products or bursts of the number of service requests. For this experiment, such \textit{quantity} drift is modeled by increasing the number of samples drawn from the data generator every 50 epochs, starting with 5,000 samples, then increasing to 10,000 and 15,000. The SLOs were set to 2 seconds for time and 97\% for model performance SLO. The number of federated clients was set to 10.
The dynamic of both SLO fulfillments is shown in Figure~\ref{fig:slos_10_clients_quantity_skew}. 

\begin{figure}[htbp]
\centerline{\includegraphics[width=\columnwidth]{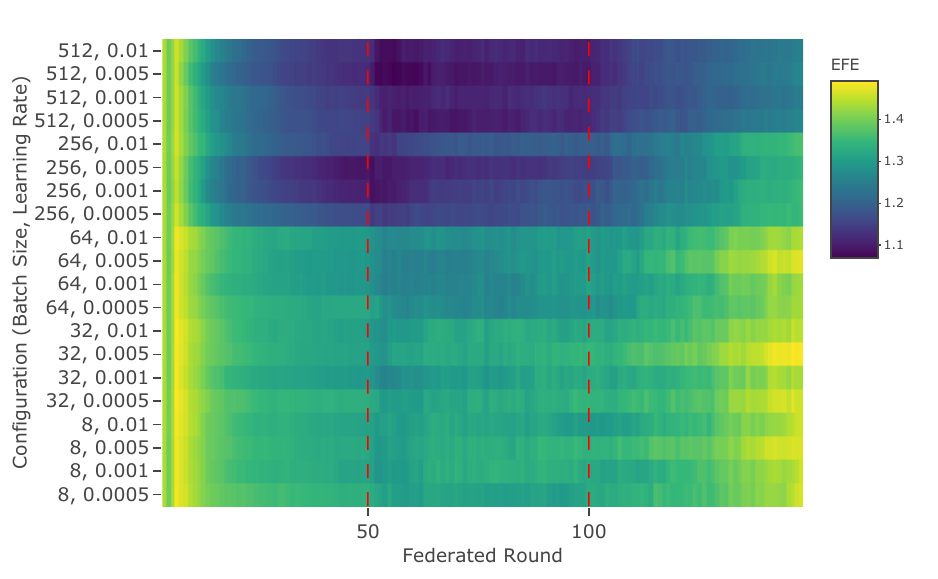}}
\caption{Mean EFE with two quantity drifts with each line representing one possible training configuration. Lower values represent configurations that the AIF agents favor.}
\label{fig:efe_10_clients_quantity}
\end{figure}

For the fixed ``optimal'' baseline, the cumulative fulfillment was high for both SLOs until the first quantity drift occurred at the 50th round. After this round, time SLO stopped being consistently fulfilled, which led to a steady decline in time SLO fulfillment. However, looking at the AIF approach, it is evident that despite the time SLO becoming more challenging to fulfill, it still manages to recover after the quantity drift, with the mean time SLO fulfillment per round being consistently larger than 50\%. Still, after the second quantity drift happened, the time SLO became even more challenging, which led to a more prominent decline in time SLO fulfillment. To understand the choices made by the AIF agents, we examine mean EFE over all configurations at each epoch (Figure~\ref{fig:efe_10_clients_quantity}). 

It is visible how the configurations preferred by the AIF agents change after the observed environmental changes. In the beginning, it is shown how agents slowly come to prefer configurations (256, 0.005) and (256, 0.001) (compared to the fixed baseline being (256, 0.01)). However, after the first quantity drift, this preference shifted to a larger batch size of 512. This is an expected behavior as the amount of data doubled, but the time constraint remained the same. Still, after the amount of data increased again, there was no more space to increase the batch size. Therefore, the increase of EFE across all configurations can be observed, signaling that. 

\begin{figure}[htbp]
\centerline{\includegraphics[width=\columnwidth]{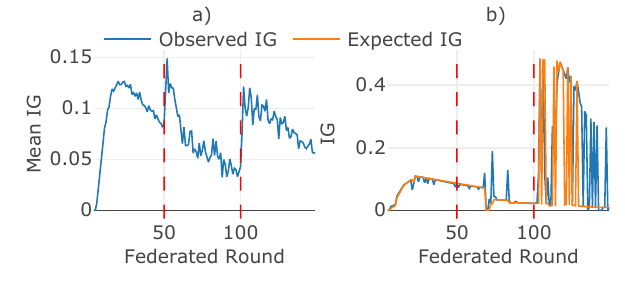}}
\caption{a) Mean observed IG over all clients and repetitions, b) Observed and expected information gain for one client at one run}
\label{fig:ig_both}
\end{figure}

As shown in Equation~\eqref{eq:efe}, the information gain term accounts for explorative behavior and is compared to the actual \textit{observed} information gain during each federated round to estimate how ``surprised'' the agent is. Figure~\ref{fig:ig_both} a shows the mean information gain across clients.

Here, the process of agents adapting to the environment can be seen, with the observed IG decreasing as the BN becomes more confident. However, two prominent spikes occur right after the quantity drifts, illustrating the ``surprised'' agents detecting the environmental change.

To better illustrate the dynamics of looking for the best configuration after the second quantity drift, the entire history of expected and observed IG of one client can be observed (Figure~\ref{fig:ig_both} b). This client initially settled down for a configuration of (512, 0,01), which worked for the agent for 68 epochs due to the observed IG being less or equal to the agent’s expectations. However, at round 68, the agent was surprised because the time SLO was not fulfilled despite using the ``time-proven'' configuration. Despite being surprised, the agent only retrained the structure of the BN (indicated by the abrupt change in the expected IG). It happened several times more, but the agent preferred exploiting its knowledge. After round 100, the agent again started to be surprised, leading to a change in strategy. The agent went exploring, as visible by the increased \textit{expected} information gain. These spikes are also associated with the agent choosing previously unexplored (or poorly explored) configurations. For instance, round 104 corresponds to the agent choosing configuration (64, 0.005). This example illustrates how AIF agents treat changes in the observed environment and can independently balance between exploration and exploitation.

\begin{figure}[htbp]
\centerline{\includegraphics[width=\columnwidth]{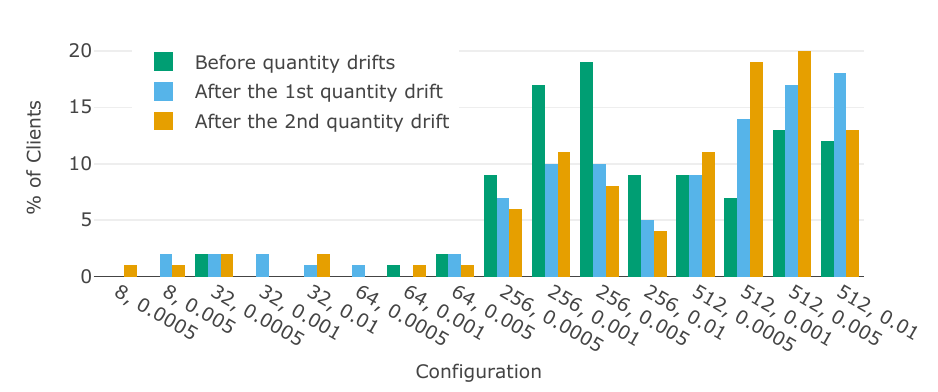}}
\caption{Chosen configurations distributions before and after quantity drifts}
\label{fig:10_quantity_dist}
\end{figure}

As the setup for the experiment consisted of ten clients and ten separate experiments, it is possible to represent configurations preferred by the agents at three critical points (before the first quantity drift, before the second quantity drift, and at the end of the experiment) into distributions over configurations. These distributions are given in Figure~\ref{fig:10_quantity_dist}. 

Despite the expected behavior, a minority of agents still settle for clearly non-optimal configurations. Another observation is the shift in the preferred configurations after the first quantity drift towards a bigger batch size. To quantify the observed changes in the distributions, Fisher's exact test was performed on the distributions with batch sizes 256 and 512 (too low values were filtered out to focus the test on the sensible configurations as the rest represent wrong configurations and are irrelevant for the test). The p-value of this test was reported to be 0.0293, showing the statistical significance in the observed changes between the distribution of the preferred configurations before and after the first quantity drift. 

Observed EFE after the second quantity drift shows that the agents' behavior is dictated by the combination of available configurations and set SLOs. To better estimate the effect SLOs have on the preferred configurations, a set of experiments was conducted that modified the SLOs considered in the experiments. Table~\ref{table:slo_experiments} shows the SLOs chosen for each experiment compared to the SLOs considered in the previous experiment. 

\begin{table}[htb]
\centering
\caption{Comparison of Time and Performance SLOs for Different Experiments}
\label{table:slo_experiments}
\resizebox{\columnwidth}{!}{%
\begin{tabular}{ccc}
\hline
Experiment               & Time SLO (s) & Performance SLO (\%) \\ \hline
\rowcolor[HTML]{EFEFEF} 
Fulfillable SLOs          & 2            & 97                     \\
Unfulfillable SLOs       & 0.1          & 99.5                   \\
\rowcolor[HTML]{EFEFEF} 
Easily Fullfillable SLOs & 3600         & 50                     \\
Time Relaxed             & 3600            & 97                     \\
\rowcolor[HTML]{EFEFEF} 
Performance Relaxed      & 2            & 50                    
\end{tabular}%
}
\end{table}

\begin{figure}[htbp]
\centerline{\includegraphics[width=\columnwidth]{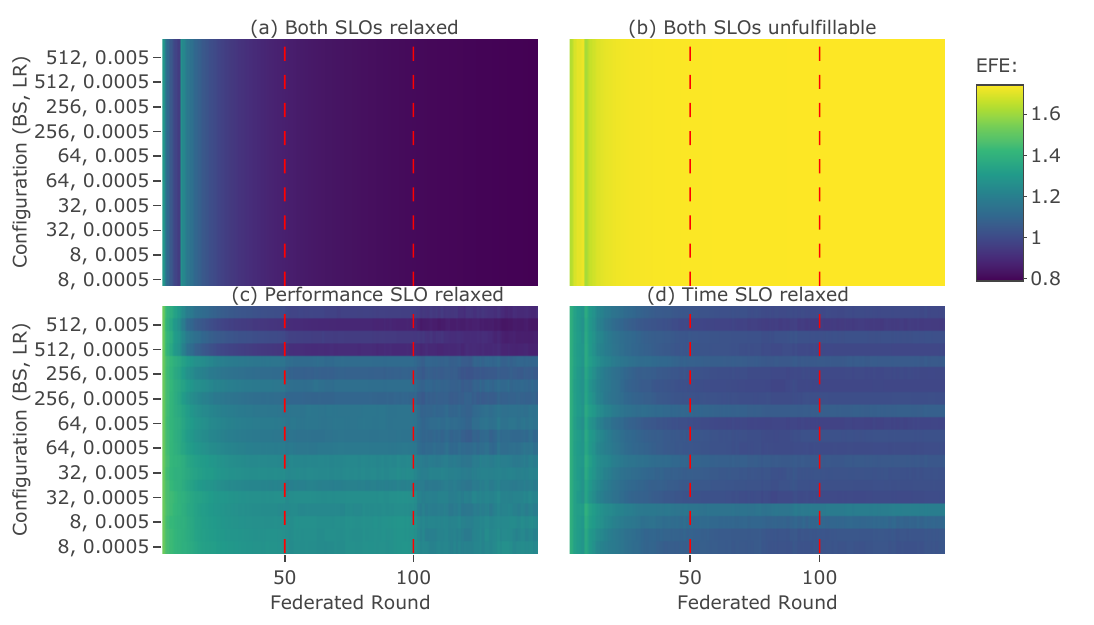}}
\caption{Mean EFE under different SLOs setups and two quantity drifts}
\label{fig:different_SLOs}
\end{figure}

Figure~\ref{fig:different_SLOs} shows the mean EFE over five clients used for experiments and ten repetitions.

The EFE shows that having not challenging or unrealistic SLOs (Figure~\ref{fig:different_SLOs} a and b, respectively) leads to EFE being consistent across all available configurations, with EFE being high when the target SLOs are unfulfillable and EFE being equally low when the targets are too easy to fulfill. The situation is different when one of the two SLOs is relaxed while the other is somehow challenging to fulfill. Thus, when performance is relaxed, the agents are incentivized to optimize the behavior towards time SLO, leading to agents settling for the largest available batch size (512) regardless of the learning rate. The situation is different regarding time-relaxed SLO – when the performance is targeted, agents tend to explore more configurations as the task is still not that challenging for them, leading to diverse behaviors. Still, the resulting EFE heatmap expresses some preference bias compared to the experiment with two SLOs relaxed. This ``uncertain'' behavior could be attributed to quantity drifts not impacting the performance goals.  

The evaluation shows the potential of AIF agents in detecting the changes in the environment and the ability to initiate system re-configuration with no human supervision. However, the SLO fulfillment is not perfect. Two main explanations were identified up until now: (1) ``unsupervised'' BN structure learning using Hill Climb Search may struggle to discover meaningful causal relationships when limited data is available~\cite{kitson_survey_2023}, (2) in the absence of observations with both SLOs fulfilled, an agent may either stuck in forever exploring state or stick to the strategy that guarantees at least one SLO fulfillment and focus on exploiting sub-optimal behavior. 

After dissecting the behaviour of the agents in a simpler scenario, the next section focuses on more practical experimental setups and presents evaluation of AIF agents' behaviour in the presence of concept drifts and resources heterogeneity.

\section{Results} \label{sec:results}

\subsection{Hyperparameter Tuning Comparison} 

To demonstrate both the validity and advantages of the proposed AIF agents, we compare them against Optuna, a state-of-the-art framework for hyperparameter optimization in ML. Optuna serves as a strong baseline due to its efficiency in finding optimal configurations. However, unlike Optuna, which focuses on static optimization, AIF agents offer continuous adaptability, making them better suited for dynamic and heterogeneous FL environments.

\subsubsection{Concept Drift}

\begin{figure}[htbp]
\centerline{\includegraphics[width=\columnwidth]{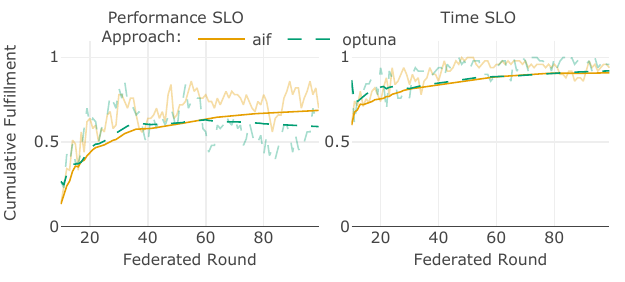}}
\caption{Mean cumulative SLOs fulfillment for the scenario with a persistent concept drift}
\label{fig:slos_concept_optuna}
\end{figure}

The first experiment included a concept drift, which was present from the beginning of the training, and both approaches needed to find the best configuration to meet the time SLO of 3s and performance SLO of 85\%. The SLOS fulfillment is shown in Figure~\ref{fig:slos_concept_optuna}.

It is seen that both Optuna and AIF have similar time SLO
fulfillment, with performance SLO being slightly better for AIF after training round 60. 

\begin{figure}[htbp]
\centerline{\includegraphics[width=\columnwidth]{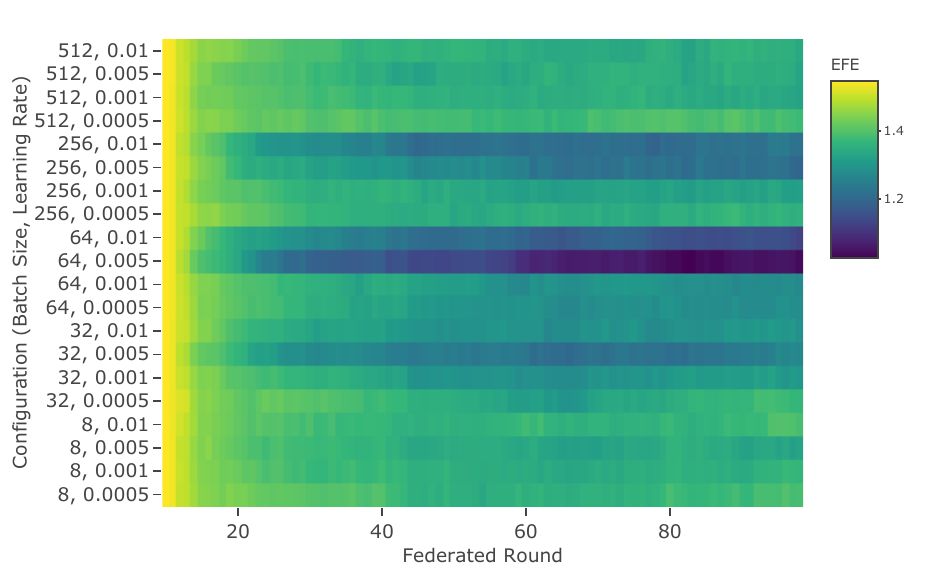}}
\caption{Mean EFE for the scenario with a persistent concept drift}
\label{fig:efe_concept_optuna}
\end{figure}

To assess what configurations were favored by both approaches, EFE (Figure~\ref{fig:efe_concept_optuna}) and normalized objectives can be examined for AIF and Optuna, respectively.

\begin{figure}[htbp]
\centerline{\includegraphics[width=\columnwidth]{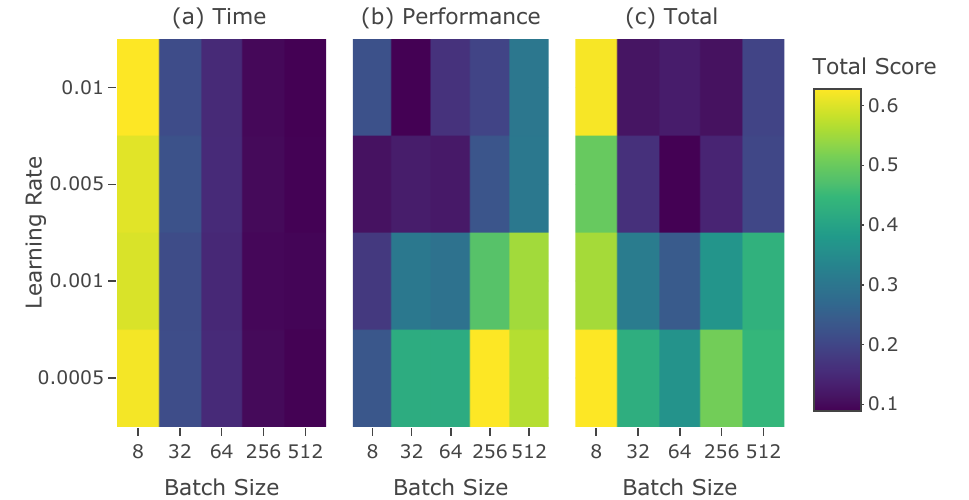}}
\caption{Optuna preferred configurations at the end (last 10 rounds) of training}
\label{fig:concept_optuna_preferences}
\end{figure}

For AIF, the best configuration based on EFE is (64, 0.005). For Optuna, the result is based on separate time and performance assessments and their combination (Figure~\ref{fig:concept_optuna_preferences}). For the time SLO, the best configurations are those of larger batch sizes, as expected. For performance, Optuna focused on a set of configurations from batch size 8 through 64 and learning rates of 0.001 through 0.01. The combination of both SLOs lands in the range of batch sizes (32, 64, 256) and learning rates of (0.005, 0.01), with (64, 0.005) being the most preferred configuration. 

Comparing the two approaches, it is seen that both AIF agents and Optuna converge to similar configurations in this scenario.

\subsubsection{Concept and Quantity Drifts after Round 50}

Next evaluation scenario introduced two drifts at the same time, simulating a change in both quantity and distribution of data received by the FL clients. 

\begin{figure}[htbp]
\centerline{\includegraphics[width=\columnwidth]{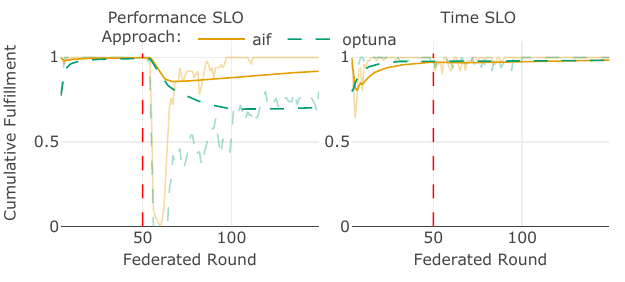}}
\caption{Mean cumulative SLOs fulfillment for the scenario with concept and quantity drifts after round 50}
\label{fig:slos_concept_quantity_optuna_2}
\end{figure}

The scenario was set as follows. First, 10,000 samples were used per FL training round with no concept drift; after round 50 -- a slight drift appeared and 20,000 samples started to be drawn from the data generator. The resulting SLOs fulfillment is presented in Figure~\ref{fig:slos_concept_quantity_optuna_2} for time SLO -- 4 seconds and performance SLO -- 90\%. 

It is seen that adding concept drift to data leads to an immediate decline in performance SLO as previously established configurations and the trained model do not account for that. However, AIF agents are able to return to a relatively high performance SLO fulfillment in a faster and more reliable way than Optuna.

\begin{figure}[htbp]
\centerline{\includegraphics[width=\columnwidth]{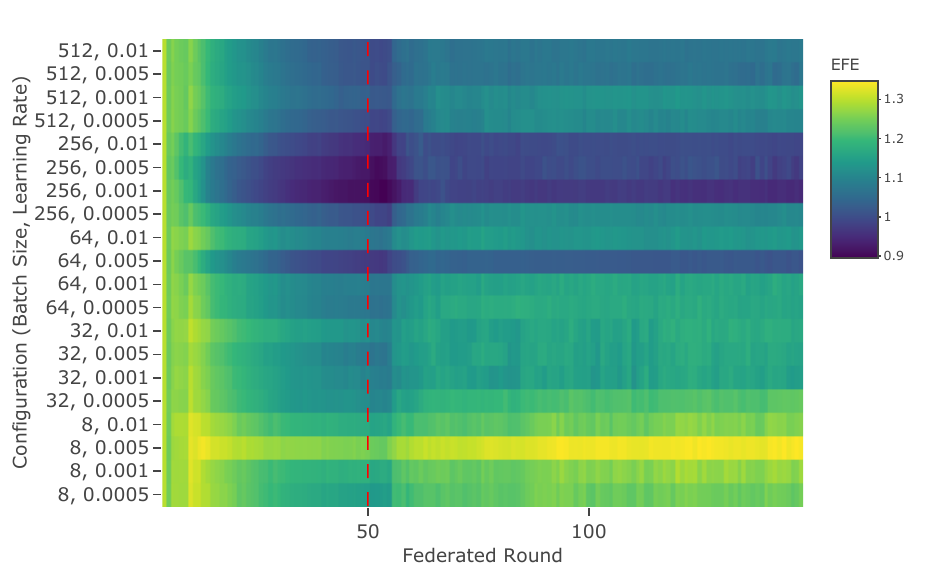}}
\caption{Mean EFE for the scenario with concept and quantity drifts after round 50} 
\label{fig:efe_concept_quantity_optuna_2}
\end{figure}

Based on EFE shown in Figure~\ref{fig:efe_concept_quantity_optuna_2}, it is seen that after the drifts appeared after round 50, a set of the previously favored configurations (256, 0.005) and (64, 0.005) gradually lost their importance and agents focused mostly on (256, 0.001).

These experiments illustrate that AIF agents align with commonly used hyperparameter optimization methods while providing adaptability where necessary. Compared to the state-of-the-art Optuna framework adapted to the lifelong heterogeneous FL scenario, AIF agents are able to fully recover from the concept drift after 43 rounds, while Optuna fails to return to its best performance even after 100 rounds after the drift introduction.

\subsubsection{Device Heterogeneity}

The next experiment focused on inspecting the ability of AIF agents to adapt to the resources available to the agents located at the edge devices and aligning them with the SLOs set. 

For the experiments, three edge devices (Raspberry Pi 5, Nvidia Orin NX, and Nvidia Xavier NX) were used as separate federated clients and were tasked to participate in the FL for 75 epochs, while a Raspberry Pi 4 device served as the FL orchestrator. No data drifts were introduced in this experiment. Due to the differences in resources, it was expected that agents would prefer different configurations under the same target SLOs. In addition to different available resources, the size of the neural networks was also changed, so to change the utilization of available resources on the device.

\begin{figure}[htbp]
\centerline{\includegraphics[width=\columnwidth]{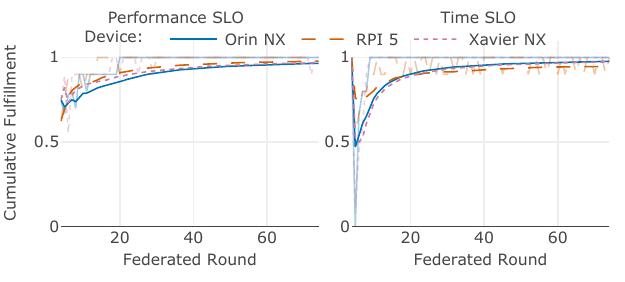}}
\caption{Mean SLOs fulfillment for different devices. Time SLO: 2 seconds, performance SLO: 97\%}
\label{fig:slos_devices}
\end{figure}

\begin{figure}[htbp]
\centerline{\includegraphics[width=\columnwidth]{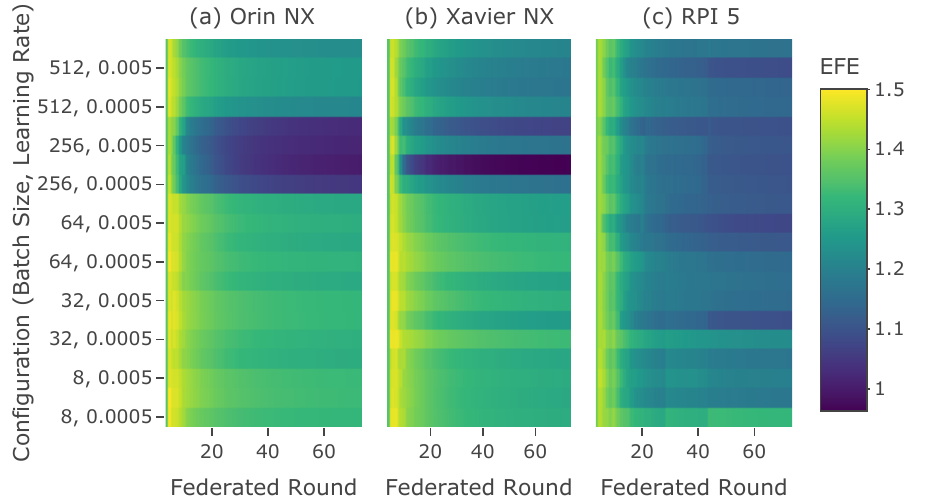}}
\caption{Mean EFE for different devices}
\label{fig:devices_efe}
\end{figure}

The first experiment featured a smaller neural network (64 and 32 units in 2 layers). The comparison of SLO fulfillment across various devices is shown in Figure~\ref{fig:slos_devices}, and mean EFE is shown in Figure~\ref{fig:devices_efe}. 

It is shown that all devices manage to fulfill the set SLOs. It is also worth noting that Raspberry Pi 5 managed to fulfill performance SLO slightly faster than devices that used GPU but occasionally struggled to maintain flawless time SLO fulfillment. When looking at mean EFE, it is clear that in order to better utilize its cores (as there is no GPU), Raspberry Pi can successfully employ a vast range of configurations, while devices with GPU choose bigger batch sizes to better utilize their parallelization capabilities. 

\begin{figure}[htbp]
\centerline{\includegraphics[width=\columnwidth]{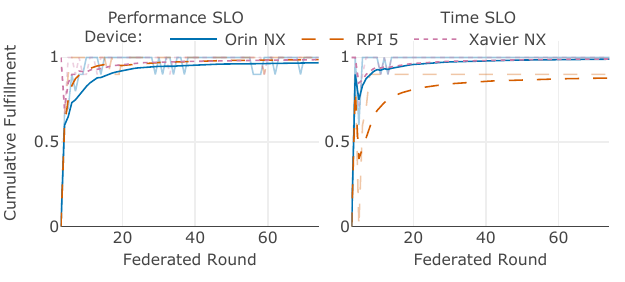}}
\caption{Mean SLOs fulfillment for different devices with a wider network. Time SLO: 15 seconds, performance SLO: 97\%}
\label{fig:slos_devices_bigger_model}
\end{figure}

\begin{figure}[htbp]
\centerline{\includegraphics[width=\columnwidth]{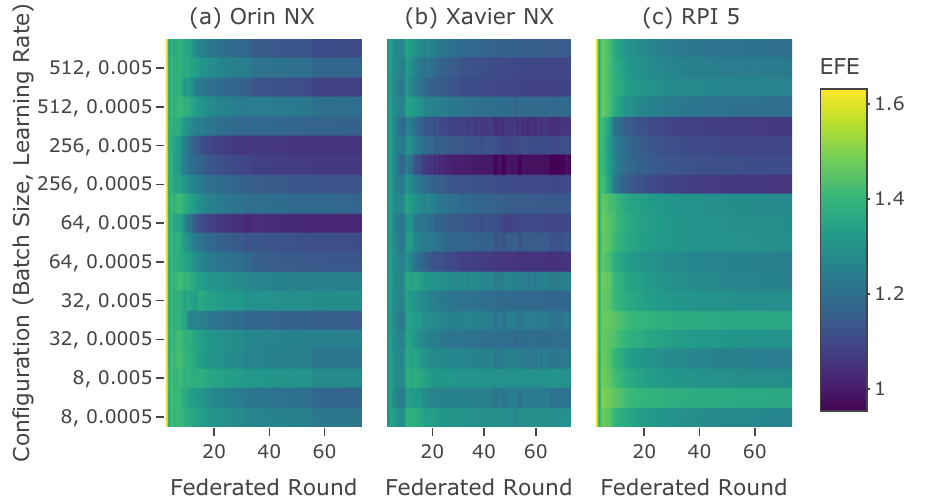}}
\caption{Mean EFE for different devices and wider network}
\label{fig:devices_efe_bigger_model}
\end{figure}

The next experiment was conducted with the same set of devices but with a wider neural network (5120 and 512 units compared to 64 and 32 used in the previous experiment). The time SLO was adjusted to 15 seconds. 
The resulting SLOs fulfillment is shown in Figure~\ref{fig:slos_devices_bigger_model} and
mean EFE is shown in Figure~\ref{fig:devices_efe_bigger_model}. The performance SLO fulfillment at the final FL round was 96.9\% for Orin NX and 98.7\% for RPI 5 and Xavier NX, while the time SLO was 99\% for Nvidia devices and 87.8\% for RPI 5. 

Here the change in the model architecture impacted the optimal configuration choice for Raspberry Pi 5. As seen from the mean EFE, the batch size had to be increased to 256 to fit into the time and performance SLOs, while Nvidia devices utilized a more comprehensive range of configurations. 

Results presented in this section show how AIF agents display adaptive behaviour in the presence of changing data as well as adaptation to the local resources to ensure the fulfillment of set SLOs.

\section{Conclusion}
This work presented AIF agents that are able to adaptively change their behavior in response to dynamic environments. We evaluated the proposed AIF agents in lifelong heterogeneous FL, utilizing a set of both dynamic data and diverse devices. We showed that AIF agents are able to fulfill competing SLOs and unfolded the behaviors of agents. We compared AIF agents to the hyperparameter-tuning framework Optuna adjusted to lifelong learning and showcased how the adaptive nature of AIF agents allows for faster performance recovery in the presence of data drift.

Future work can further expand the usage of the active inference framework to orchestrate distributed learning systems, for instance, by fulfilling system-level SLOs, such as fairness of participation or global model performance. Another line of research can target the scalability of the proposed framework, as the usage of BNs can potentially introduce computational bottleneck as the number of considered vertices and their cardinalities grow.
Enhancements of the current method can improve the ability of the agents to find causal dependencies in limited data, making them more robust, targeting the limitations of the discrete BN, introducing temporal dependencies to capture the environmental dynamics more precisely, and providing more nuanced SLOs specifications to enable tracking SLOs in a range. 
\balance
\section*{Acknowledgments}
We thank Alexander Knoll for providing us with the hardware infrastructure and Pantelis Frangoudis for his valuable suggestions and feedback.

The work of Anastasiya Danilenka was conducted during the research visit funded by the Warsaw University of Technology within the Excellence Initiative: Research University (IDUB) programme. The work of Maria Ganzha was co-funded by the Centre for Priority Research Area Artificial Intelligence and Robotics of Warsaw University of Technology within the Excellence Initiative: Research University (IDUB) programme. Further, this work was supported in part by EU Horizon under Grants 101135576 (INTEND), 101079214 (AIoTwin), and 101070186 (TEADAL). Ayuda CNS2023-144359 financiada por MICIU/AEI/10.13039/501100011033 y por la Unión Europea NextGenerationEU/PRTR.

\bibliographystyle{IEEEtran}
\bibliography{bibliography}
\newpage

\vfill

\end{document}